\documentclass[sigconf]{acmart} 

\AtBeginDocument{%
  }

\copyrightyear{2023}
\acmYear{2023}
\setcopyright{acmlicensed}
\acmConference[MM '23] {Proceedings of the 31st ACM International Conference on Multimedia}{October 29--November 3, 2023}{Ottawa, ON, Canada.}
\acmBooktitle{Proceedings of the 31st ACM International Conference on Multimedia (MM '23), October 29--November 3, 2023, Ottawa, ON, Canada}
\acmPrice{15.00}
\acmISBN{979-8-4007-0108-5/23/10}
\acmDOI{10.1145/3581783.3612273}

\usepackage{bbding}
\usepackage{float}

\begin{document}

\title{Exploiting Low-confidence Pseudo-labels for Source-free \\Object Detection}

	\author{Zhihong Chen}
	\email{zhchen@mail.ustc.edu.cn}
	\orcid{0009-0005-4162-0681}
	\affiliation{%
		\institution{University of Science and Technology of China}
		\city{Hefei}
		\country{China}
	}
	
	\author{Zilei Wang}
	\authornote{Corresponding Author}
	\email{zlwang@ustc.edu.cn}
	\orcid{0000-0003-1822-3731}
	\affiliation{%
		\institution{University of Science and Technology of China}
		\city{Hefei}
		\country{China}
	}

	\author{Yixin Zhang}
	\email{zhyx12@mail.ustc.edu.cn}
	\orcid{0000-0002-4513-1106}
	\affiliation{%
		\institution{University of Science and Technology of China}
		\city{Hefei}
		\country{China}
	}

\renewcommand{\shortauthors}{Zhihong Chen, Zilei Wang, and Yixin Zhang}

\begin{abstract}
Source-free object detection (SFOD) aims to adapt a source-trained detector to an unlabeled target domain without access to the labeled source data. Current SFOD methods utilize a threshold-based pseudo-label approach in the adaptation phase, which is typically limited to high-confidence pseudo-labels and results in a loss of information. To address this issue, we propose a new approach to take full advantage of pseudo-labels by introducing high and low confidence thresholds. Specifically, the pseudo-labels with confidence scores above the high threshold are used conventionally, while those between the low and high thresholds are exploited using the Low-confidence Pseudo-labels Utilization (LPU) module. The LPU module consists of Proposal Soft Training (PST) and Local Spatial Contrastive Learning (LSCL). PST generates soft labels of proposals for soft training, which can mitigate the label mismatch problem. LSCL exploits the local spatial relationship of proposals to improve the model's ability to differentiate between spatially adjacent proposals, thereby optimizing representational features further. Combining the two components overcomes the challenges faced by traditional methods in utilizing low-confidence pseudo-labels. Extensive experiments on five cross-domain object detection benchmarks demonstrate that our proposed method  outperforms the previous SFOD methods, achieving state-of-the-art performance.
\end{abstract}




\maketitle

\section{Introduction} \label{introduction}
\begin{figure*}[h]
  \centering
  \includegraphics[width=\linewidth]{ 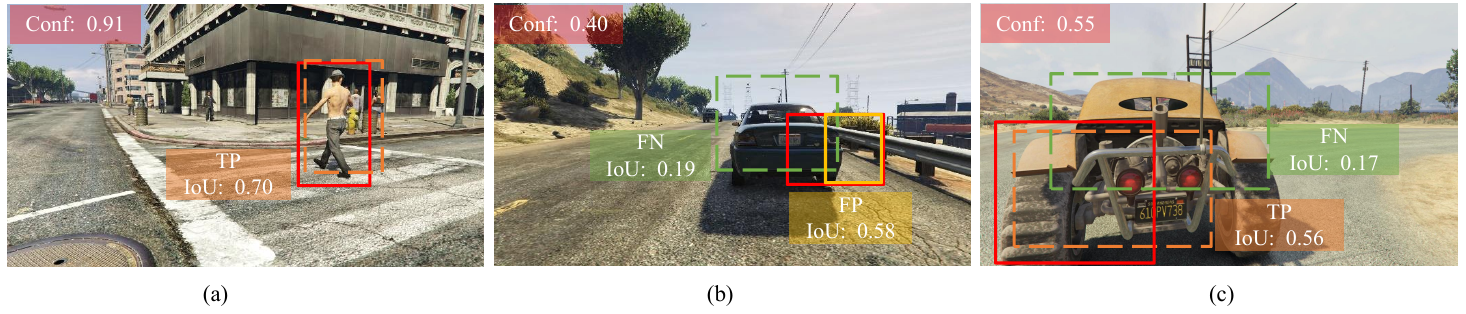}
  \caption{Demonstrative cases of the IoU-based label assignment. The red solid bounding boxes correspond to the pseudo-bbox, and their confidence level is indicated by 'Conf.' Meanwhile, the dashed bounding boxes represent generated proposals with different colors used. The intersection over union (IoU) value between the proposal and its corresponding pseudo-bbox is shown as 'IoU'. 'TP' indicates the foreground label is correctly assigned to the proposal. 'FP' denotes an incorrect assignment of the foreground label to the proposal, which should have been assigned a background label instead. Similarly, 'FN' indicates when the background label is wrongly assigned to the proposal, which should have been labeled as foreground.} \label{fig:mismatch}
\end{figure*}

\begin{figure}[h]
  \centering
  \includegraphics[width=\linewidth]{ 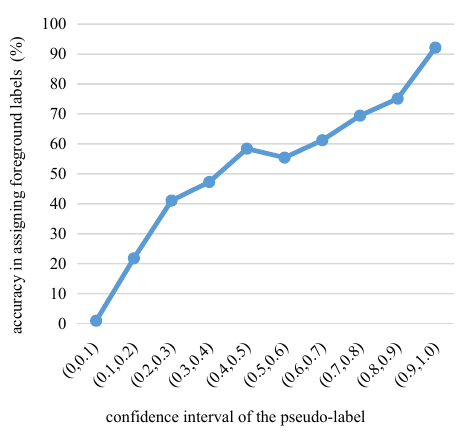}
  \caption{Accuracy in assigning foreground labels for different confidence intervals of the pseudo-label. It is obtained using 500 randomly selected images from Cityscapes~\cite{cityscapes}.} \label{fig:acc_pseudo}
\end{figure}

Deep convolutional neural networks have led to significant advancements in image object detection (\textit{e.g.}, Faster R-CNN~\cite{faster_rcnn} and YOLO~\cite{yolo}). However, when a target detector encounters a novel environment with domain shift~\cite{da_faster}, such as changes in weather or style, its performance may significantly deteriorate. To tackle this challenge, unsupervised domain adaptive object detection (UDA-OD) has received extensive research attention in recent years~\cite{sw_faster,wu2021vector, Zhang_2021_CVPR}.
The UDA approach assumes access to data from both the source and target domains. However, in practice, the source domain data may be inaccessible due to data privacy, distributed storage, or inconvenient data transfer. As a result, source-free object detection (SFOD)~\cite{free_lunch,lods,a2sfod} has emerged as a hot topic. SFOD involves using only a pre-trained model on the source domain and unlabeled data in the target domain without requiring labeled source data.

Most SFOD methods employ the mean-teacher framework for training, given the absence of manually labeled data. This structure involves two main components: the teacher and the student. The teacher model guides the learning of the student model. Typically, a threshold-based pseudo-labeling approach is applied, where only pseudo-labels with confidence scores above a certain threshold are used for training. However, existing methods often set a very high threshold (\textit{e.g.}, 0.8) empirically to ensure high-quality pseudo-labeling. Moreover, due to the class imbalance in data distribution, the optimal threshold may vary for different categories \cite{zhang2021flexmatch}. Consequently, the traditional settings lead to discarding valuable information by neglecting the low-confidence samples.

This motivates us to develop a method to utilize pseudo-labels more effectively, especially for those with low confidence. We discovered that the main obstacle in leveraging low-confidence pseudo-labels lies in the rough IoU-based label assignment used in the traditional approach \cite{faster_rcnn}. Specifically, in the conventional method, a proposal is labeled with the label of a pseudo-bounding box if its IoU with that pseudo-bbox is greater than a certain threshold. Otherwise, it is considered as the background class. Figure \ref{fig:mismatch} illustrates a few representative examples. From Figure \ref{fig:mismatch}(a) and Figure \ref{fig:mismatch}(b), it is evident that proposals tend to receive more accurate labels when the confidence level of a pseudo-label is high. However, when the confidence level of a pseudo-label is low, the inaccurate position of the bounding box (bbox) can easily lead to label misassignment. We conducted a quantitative analysis of this problem, and the results are shown in Figure \ref{fig:acc_pseudo}. The results clearly indicate that label misassignment becomes more pronounced as the confidence of a pseudo-label decreases. Hence, the conventional label assignment approach is unsuitable for exploiting low-confidence pseudo-labels.

To overcome this problem, we propose a novel approach called Low-confidence Pseudo-labels Utilization (LPU). Our method sets two thresholds, a high threshold $\theta_h$, and a low threshold $\theta_l$. We use the conventional approach and directly utilize pseudo-labels with confidence scores above $\theta_h$, as these pseudo-labels are considered sufficiently accurate. However, for pseudo-labeled data with confidence scores between the low and high thresholds, we employ our proposed LPU module for training. The LPU module comprises Proposal Soft Training (PST) and Local Spatial Contrastive Learning (LSCL).

The objective of PST is to assign more accurate labels to proposals. Specifically, we feed the proposals generated by the student model into the teacher model, extract their features, use the class predictions generated after the RCNN classification layer as soft labels, and then perform self-training. Unlike the traditional IoU-based label assignment, where hard labels may introduce noise due to the low confidence of pseudo-labels, our approach employs soft labels, which have two main advantages. Firstly, soft labels preserve intra-class and inter-class associations and carry more information, resulting in a stronger generalization ability for the model and increased robustness to noise. Secondly, the teacher model updates its parameters using Exponential Moving Average (EMA), which updates the parameters more slowly than standard training methods, preserving the source model's parameter information and preventing the forgetting of source hypotheses during training. Overall, PST improves the label assignment accuracy of the student model by leveraging the more accurate and reliable soft labels generated by the teacher model. This approach addresses the label mismatch issue, enabling the model to capture semantic information better and improve performance.

Additionally, to improve the model's ability to differentiate between neighboring proposals in spatial proximity, we introduce LSCL. In Figure \ref{fig:mismatch}(c), the red solid line represents the pseudo-bbox with low confidence, while the dashed line indicates the two adjacent proposals. If we use the traditional IoU-based label assignment method, the green proposal will be mistakenly labeled as the background class. However, we can observe that the orange proposal is correctly labeled and located near the green proposal in space. This motivates us to improve the feature representation of proposals by exploiting the local spatial relationships among them. To achieve this, LSCL employs an IoU-mixup approach to merge the proposals generated by the student and teacher models. The merged proposals are then optimized using the adjacent proposal contrastive-consistency loss. This encourages the model to explore finer-grained cues between neighboring proposals and form more robust classification boundaries, improving performance.

The contributions of this work are summarized as follows: 
\begin{itemize}
\item We propose a novel LPU approach that addresses the low utilization of pseudo-labels in the conventional threshold-based pseudo-labeling approach. The LPU module effectively leverages the informative content of low-confidence pseudo-labels.
\item Within the LPU module, we apply PST to mitigate the label mismatch resulting from the IoU-based label assignment method. Moreover, LSCL helps the model better understand the relationship between neighboring proposals and learn more accurate and robust representations.
\item The effectiveness of the proposed method is evaluated across four SFOD tasks on five detection datasets. Our method outperforms existing source-free domain adaptation methods and many unsupervised domain adaptation methods.
\end{itemize}


\section{RELATED WORK}

\subsection{Unsupervised domain adaptive object detection}
Unsupervised domain adaptive object detection (UDAOD) aims to address the domain shift challenges in object detection tasks. Existing UDAOD methods can be broadly categorized into three groups. The first category is based on adversarial feature learning, as demonstrated in \cite{da_faster,sw_faster,chen2020harmonizing,categorical_da,wu2021vector,chen2021dual}. This adaptation approach trains object detector models adversarially with the help of a domain discriminator. Specifically, the detector model is trained to generate features that can deceive the domain discriminator, whose task is to correctly classify these into source and target domains. The second category employs a self-training strategy, as shown in \cite{inoue2018cross,kim2019self,khodabandeh2019robust,zhao2020collaborative}. These methods use the source-trained detector to generate high-quality pseudo-labels on the target domain. The third category is image-to-image translation, as illustrated in \cite{cai2019exploring,categorical_da,hsu2020progressive,chen2020harmonizing,shen2021cdtd}. This strategy employs an image translation model to convert the target image into a source-like image or vice versa. This mitigates the differences in the distribution of source and target domains, thus facilitating adaptation. Although these methods achieve good performance, all of the above domain adaptation methods require access to the source domain data during target adaptation.

\subsection{Source-free domain adaptation}
Recently, many methods \cite{kundu2020towards,kundu2020universal,ahmed2021unsupervised,hou2021visualizing,xia2021adaptive,yang2021generalized,kundu2022balancing,kundu2022concurrent,ding2022source} have emerged for solving Source-Free Domain Adaptation(SFDA), which aims to adapt a detector pre-trained on the source domain to an unlabeled target domain without source data. Liang \textit{et al.} \cite{shot} uses pseudo labeling and information maximization to match target features with a fixed source classifier. Li \textit{et al.} \cite{3CGAN} synthesizes labeled target domain training images based on a conditional GAN as a way to provide supervision for adaptation. Yang \textit{et al.} \cite{NRC} proposes neighborhood clustering, which performs predictive consistency among local neighborhoods.

\begin{figure*}[h]
  \centering
  \includegraphics[width=\linewidth]{ 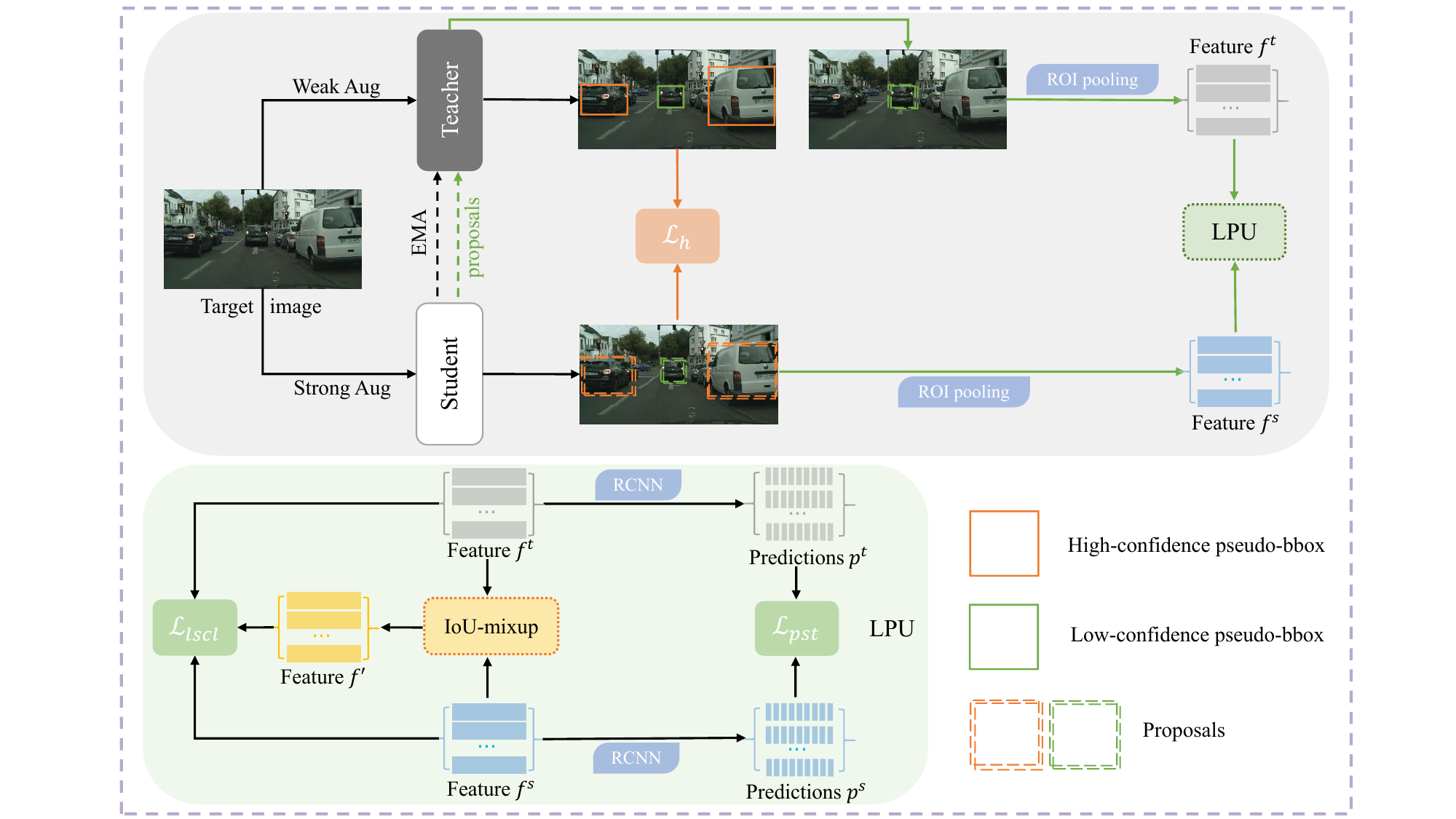}
  \caption{The framework of our method. We follow a mean teacher framework for training the detector model. High-confidence pseudo-labels are used conventionally, while low-confidence pseudo-labels are exploited using the LPU module. The loss function $\mathcal{L}_{pst}$ corresponds to Proposal Soft training (PST), which generates soft labels for proposal training. On the other hand, the loss function $\mathcal{L}_{lscl}$ corresponds to Local Spatial Contrastive Learning (LSCL), which leverages the local spatial relationship to optimize the feature representation of proposals.} 
  \label{method}
\end{figure*}

Due to the complex background and negative examples, there are few methods for source-free object detection (SFOD). Li \textit{et al.} \cite{free_lunch} suggests a self-entropy descent policy for determining a suitable confidence threshold and conducting self-training with generated pseudo-labels. Li \textit{et al.} \cite{lods} implements domain adaptation by allowing the detector to learn to ignore domain styles. Chu \textit{et al.} \cite{a2sfod} divides the target dataset into source-similar and source-dissimilar parts and aligns them in the feature space by adversarial learning. \cite{IRG} designs an instance relation graph network and combines it with contrastive learning to transfer knowledge to the target domain. However, these methods do not effectively utilize the information provided by pseudo-labels. Even though \cite{free_lunch} proposes a way to find a suitable confidence threshold, a single threshold alone cannot address the problem discussed in the third paragraph of the section \ref{introduction}.

\section{METHOD}

\subsection{Preliminaries}
\subsubsection{Problem statement}
Suppose the labeled source domain $D_{s}=\left\{\left(x_{s}^{i}, y_{s}^{i}\right)\right\}_{i=1}^{N_{s}}$, where $x_{s}^{i}$ denotes the $i^{th}$ source image and $y_{s}^{i}$ denotes the corresponding ground-truth, $N_{s}$ denotes the total number of source images. Target domain $D_{t}=\left\{x_{t}^{i}\right\}_{i=1}^{N_{t}}$, where $x_{t}^{i}$ denotes the $i^{th}$ target image and $N_{t}$ denotes the total number of target images. Source domain and target domain obey the same distribution. Our goal is to adapt the source model to the target domain without access to the source dataset, \textit{i.e.}, only the source model $\Theta_{s}$ and the unlabeled target data $D_{t}$ are available.

\subsubsection{Mean Teacher Framework}
Due to the unavailability of the source data, we build our approach based on the mean-teacher framework. This framework comprises two components: the teacher and student models. Both networks are initialized with the source-trained model at the start of the training process. The teacher model generates pseudo-labels for the weakly enhanced unlabeled data, whereas the student model is trained on the strongly enhanced unlabeled data using the generated pseudo-labels. As the training progresses, the parameters of the student model are updated through gradient descent. In contrast, the parameters of the teacher model are updated via an Exponential Moving Average (EMA) strategy from the student. Formally, this can be expressed as follows:
\begin{equation}
    \mathcal{L}_{mt}\left(x_{i}\right)=\mathcal{L}_{r p n}\left(x_{i}, \hat{y}_{i}\right)+\mathcal{L}_{roi}\left(x_{i}, \hat{y}_{i}\right), \label{equ_1}
\end{equation}
\begin{equation}
      \Theta_{s} \leftarrow \Theta_{s}+\gamma \frac{\partial\left(\mathcal{L}_{mt}\right)}{\partial \Theta_{s}}, \label{equ_2} 
\end{equation}
\begin{equation}
    \Theta_{t} \leftarrow \alpha \Theta_{t}+(1-\alpha) \Theta_{s}, \label{equ_3}
\end{equation}
where $x_{i}$ represents the $i^{th}$ unlabeled target image, while $\hat{y}_{i}$ represents the corresponding pseudo-label. The parameters of the source model and target model are denoted by $\Theta_{s}$ and $\Theta_{t}$, respectively. The learning rate of the student model is denoted by $\gamma$, and the teacher EMA rate is denoted by $\alpha$. Despite the mean-teacher framework's effectiveness in distilling knowledge from a source-trained model, it does not optimally utilize pseudo-label information, as discussed earlier. To address this issue, we propose a module called LPU, which aims to utilize the available pseudo-label information efficiently.

\subsection{Proposed Method}
\subsubsection{Overview}
In contrast to prior work, our approach effectively and thoroughly utilizes the pseudo-labels generated by the teacher. Specifically, we establish two confidence thresholds, a high threshold $\sigma_{h}$ (0.8 in our experiments) and a low threshold $\sigma_{l}$ (0.1 in our experiments). For the pseudo-labels with confidence exceeding the high threshold $\sigma_{h}$, we directly train them using the conventional mean-teacher framework. The training objective for this part can be expressed as follows:
\begin{equation}
    \mathcal{L}_{h}=\mathcal{L}_{r p n}\left(x_{i}, \hat{y}_{i}^{h}\right)+\mathcal{L}_{roi}\left(x_{i}, \hat{y}_{i}^{h}\right), \label{equ_4}
\end{equation}
where $\hat{y}_{i}^{h}$ represents the high-confidence pseudo-label.

For data with confidence levels between the low threshold $\sigma_{l}$ and the high threshold $\sigma_{h}$, we employ the LPU module for training to exploit the informative content of the pseudo-labels fully.  The LPU module consists of Proposal Soft Training (PST) and Local Spatial Contrastive Learning (LSCL).  The PST module utilizes the teacher model to provide more accurate and reliable soft labels for the proposals generated by the student during self-training. On the other hand, LSCL conducts an IoU-mixup on the proposals in the spatial location neighborhood, enabling the model better to understand the relationship between neighboring proposals through contrastive learning. This approach facilitates more robust and accurate learning, as PST and LSCL complement each other, effectively extracting valid information from the low-confidence pseudo-labels. Without access to the source domain data, our method efficiently utilizes the target domain for self-training and adaptation. The entire training process is depicted in Figure \ref{method}.

\begin{figure*}
    \centering
    \includegraphics{ 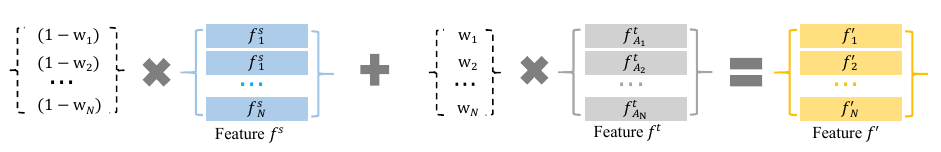}
    \caption{Illustration of IoU-mixup.}
    \label{fig:mixup}
\end{figure*}

\begin{figure}
    \centering
    \includegraphics{ 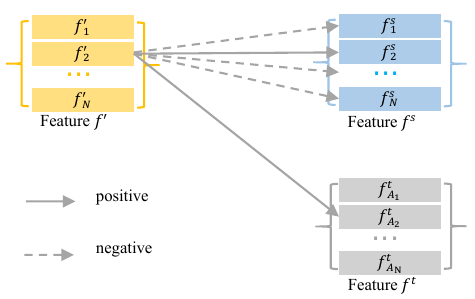}
    \caption{Illustration of contrastive consistency of adjacent proposals. }
    \label{fig:contrastive}
\end{figure}

\subsubsection{Proposal Soft Training}
As depicted in Figure \ref{fig:mismatch} and Figure \ref{fig:acc_pseudo}, the conventional IOU-based label assignment manner is susceptible to label misassignment, particularly for low-confidence pseudo-labels. Inaccurate label assignment may cause the model to update erroneously, thus adversely affecting the performance of model (as demonstrated in the experimental section). To fully capitalize on the value of low-confidence pseudo-labels, we propose PST as an alternative to the IOU-based label assignment manner. Specifically, we input proposals matching low-confidence pseudo-labels to the teacher model and leverage the class predictions output by the RCNN classification layer as the soft labels. Subsequently, proposals generated by the student model are self-trained using the soft labels provided by the teacher model based on Eq. \ref{equ_5}:
\begin{equation}
    {\mathcal{L}}_{pst}=\frac{1}{N_{p}}\sum_{i=1}^{N_{p}} \sum_{c=1}^{N_{c}}-p_{i, c}^{t} \log p_{i, c}^{s}, \label{equ_5}
\end{equation}
where $N_{p}$ refers to the total number of proposals matching the low-confidence pseudo-labels, $N_{c}$ is the number of categories, $p_{i, c}^{s}$ represents the class predictions of the $c^{th}$ category in the $i^{th}$ proposal from the student, and $p_{i, c}^{t}$ denotes the soft label corresponding to $p_{i, c}^{s}$  generated by the teacher.

There are other methods that also apply soft labels for training. \cite{soft_teacher} and \cite{scmt} simply use the logit output generated by teacher to re-weight the unsupervised loss. They essentially use the traditional IoU-based label assignment manner, the drawbacks of which we have already described earlier. \cite{label_matching} uses soft labels refined from the ROIHead of teacher to avoid NMS confusion (some proposals tend to be the same in classification score but different in localization after being refined by ROIHead). As for our PST module, we generate soft labels of proposals for soft training, which can mitigate the label mismatch problem caused by the traditional IoU-based label assignment manner. It has significant differences from all three of the above methods.

\subsubsection{Local Spatial Contrastive Learning} 
This section addresses a critical issue that complements the PST module: local prediction stability, which needs to be adequately addressed in previous SFOD work. To better understand the relationship between neighboring proposals and enhance the accuracy and robustness of the proposals' representation features, we introduce the LSCL module. The LSCL module consists of two steps: IoU-mixup and adjacent proposal contrastive-consistency.

\noindent \textbf{IoU-mixup. } We denote the ${N_{p}}$ student-generated proposals that match the low-confidence pseudo-labels as $\left\{\boldsymbol{T}_{n}\right\}_{n=1}^{N_{p}}$. The proposals will  transform into the corresponding features after passing through the ROI Pooling layer of the student or teacher, denoted as $\left\{\boldsymbol{f}_{n}^{s}\right\}_{n=1}^{N_{p}}$ and $\left\{\boldsymbol{f}_{n}^{t}\right\}_{n=1}^{N_{p}}$, respectively. For $\left\{\boldsymbol{T}_{n}\right\}_{n=1}^{N_{p}}$, we use $\left\{\boldsymbol{A}_{n}\right\}_{n=1}^{N_{p}}$ to denote the subscript corresponding to the proposal that is closest to it in spatial location, \textit{i.e.}, the one that satisfies the following constraint:

\begin{equation}
    IoU(T_{i},T_{A_i})\geq IoU(T_{i},T_{j}) \quad \forall j\in\left \{{1,..,N_p}\right \} \wedge T_{i} \neq T_{A_i}, \label{equ_6}
\end{equation}
where $\textit{IoU}$ ($T_{i}$,$T_{A_i}$) means the IoU of the proposal $T_{i}$ and $T_{A_i}$.
Then, we can initiate the IoU-mixup operation:
\begin{equation}
    w_i=IoU(T_{i},T_{A_i}), \label{equ_7}
\end{equation}
\begin{equation}
    \boldsymbol{f}_{i}^{\prime} = (1-w_i) \boldsymbol{f}_{i}^{s} + w_i \boldsymbol{f}_{A_i}^{t} \quad \forall i \in \left \{{1,..,N_p}\right \}, \label{equ_8}
\end{equation}
The process is illustrated in Figure \ref{fig:mixup}, where we perform a mixup operation on the features that correspond to the two proposals with the highest IoU score to generate an enhanced version $\left\{\boldsymbol{f}_{n}^{\prime}\right\}_{n=1}^{N_{p}}$. In Equation \ref{equ_8}, it should be noted that the mixup with ${f}_{i}^{s}$ is performed with ${f}_{A_i}^{t}$ instead of ${f}_{A_i}^{s}$. Since the inputs for the student and teacher models are different enhanced versions of the images, thus using this approach for mixup can provide more benefits to the model in tapping into these differences during subsequent contrastive learning.

\noindent \textbf{Adjacent proposal contrastive-consistency. }
As shown in Figure \ref{fig:contrastive}, taking each enhanced feature ${f_i^\prime}$ as query, we treat $f_{i}^s$ and $f_{A_i}^t$ as positive keys, and the rest of the features in $\left\{\boldsymbol{f}_{n}^{s}\right\}_{n=1}^{N_{p}}$ as negative keys, so as to construct a contrastive loss:
\begin{equation}
    \begin{aligned}
\mathcal{L}_{\text {lscl}}=-\frac{1}{N_p} & \sum_{i=1}^{N_p} \left(1-w_{i}\right) \log \frac{\exp \left(\boldsymbol{f}_{i}^{\prime \top} \boldsymbol{f}_{i}^{s} / \tau\right)}{\sum_{j=1}^{N_p} \exp \left(\boldsymbol{f}_{i}^{\prime \top}{ } \boldsymbol{f}_{j}^s / \tau\right)}+ \\
& w_i \log \frac{\exp \left(\boldsymbol{f}^{\prime \top}_{i} \boldsymbol{f}_{A_i}^{t} / \tau\right)}{\sum_{j=1}^{N_p} \exp \left(\boldsymbol{f}_{i}^{\prime}{ }^{\top} \boldsymbol{f}_{j}^s / \tau\right)},
\end{aligned}
\end{equation}
where $\tau$ is the temperature coefficient.

The module uses contrastive loss $\mathcal{L}_{\text {lscl}}$ on the features produced after IoU-mixup to learn the proximity of adjacent proposals and to identify the relative similarity between neighboring proposals and other proposals. This process enables the features to gradually capture the essential nuances, improving fine-grained discrimination and mitigating the previously mentioned label mismatch issue.

\subsubsection{Overall Loss}
The overall objective of our proposed end-to-end SFOD method is formulated as:
\begin{equation}
    \mathcal{L}_{S F OD}=\mathcal{L}_{h}+\lambda_{1}\mathcal{L}_{pst}+\lambda_{2}\mathcal{L}_{lscl}, \label{equ_10}
\end{equation} 
where ${\lambda_{1}}$ and $\lambda_{2}$ are hyperparameters to balance loss components.

\section{EXPERIMENTS}
\begin{table*}
  \caption{Experimental results of adaptation from clear to foggy dataset, \textit{i.e.}, from Cityscapes to Foggy Cityscapes.}
  \label{tab:c2f}
  \begin{tabular}{@{}ccccccccccc@{}}
    \toprule
    Methods  &Source-free      & truck         & car           & rider         & person        & train         & motor         & bicycle       & bus           & mAP           \\ \midrule
    Source only  & \XSolidBrush   & 13.9          & 36.5          & 36.7          & 29.7          & 5.0           & 20.1          & 32.7          & 30.7          & 25.7          \\ \midrule
    DA-Faster (\cite{da_faster}, CVPR 2018)  & \XSolidBrush     & 22.1          & 40.5          & 31.0          & 25.0          & 20.2          & 20.0          & 27.1          & 33.1          & 27.6          \\
    Selective DA (\cite{selectiveDA}, CVPR 2019)  & \XSolidBrush  & 26.5          & 48.5          & 38.0          & 33.5          & 23.3          & 28.0          & 33.6          & 39.0          & 33.8          \\
    SW-Faster (\cite{sw_faster}, CVPR 2019)  & \XSolidBrush     & 23.7          & 47.3          & 42.2          & 32.3          & 27.8          & 28.3          & 35.4          & 41.3          & 34.8          \\
    Categorical DA (\cite{categorical_da}, CVPR 2020) & \XSolidBrush   & 27.2          & 49.2          & 43.8          & 32.9          & 36.4          & 30.3          & 34.6          & 45.1          & 37.4          \\
    AT-Faster (\cite{at_faster}, ECCV 2020)   & \XSolidBrush     & 23.7          & 50.0          & 47.0          & 34.6          & 38.7          & 33.4          & 38.8          & 43.3          & 38.7          \\
    MeGA CDA (\cite{mega_cda}, WACV 2020)  & \XSolidBrush       & 25.4          & 52.4          & 49.0          & 37.7          & 46.9          & 34.5          & 39.0          & 49.2          & 41.8          \\
    Unbiased DA (\cite{unbiased_da}, CVPR 2021) & \XSolidBrush     & 30.0          & 49.8          & 47.3          & 33.8          & 42.1          & 33.0          & 37.3          & 48.2          & 40.4          \\ 
    TIA (\cite{TIA}, CVPR 2022)    & \XSolidBrush        & 31.1          & 49.7          & 46.3         & 34.8          & 48.6          & 37.7          & 38.1          & 52.1          & 42.3          \\ \midrule
    SFOD-Mosaic (\cite{free_lunch}, AAAI 2021)  & \Checkmark    & 25.5          & 44.5          & 40.7          & 33.2          & 22.2          & 28.4          & 34.1          & 39.0          & 33.5          \\
    LODS (\cite{lods}, CVPR 2022)   & \Checkmark        & 27.3          & 48.8          & 45.7          & 34.0          & 19.6          & \textbf{33.2} & 37.8          & 39.7          & 35.8          \\
    $A^{2}SFOD$ (\cite{a2sfod}, AAAI 2023)  & \Checkmark       & \textbf{28.1} & 44.6          & 44.1          & 32.3          & \textbf{29.0} & 31.8          & 38.9          & 34.3          & 35.4          \\
    IRG (\cite{IRG}, CVPR 2023)   & \Checkmark         & 24.4          & 51.9          & 45.2          & 37.4          & 25.2          & 31.5          & 41.6          & 39.6          & 37.1         \\
    Ours   & \Checkmark        & 24.0          & \textbf{55.4} & \textbf{50.3} & \textbf{39.0} & 21.2          & 30.3          & \textbf{44.2} & \textbf{46.0} & \textbf{38.8} \\ 
    \bottomrule
  \end{tabular}
\end{table*}

\begin{table}
  \caption{Experimental results of adaptation to a new sense, \textit{i.e.}, from KITTI to Cityscapes.}
  \label{tab:k2c}
    \begin{tabular}{@{}ccc@{}}
        \toprule
        Methods     & Source-free & AP of car     \\ \midrule
        Source only & \XSolidBrush       & 35.9          \\ \midrule
        DA-Faster (\cite{da_faster}, CVPR 2018)   & \XSolidBrush       & 38.5          \\
        SW-Faster (\cite{sw_faster}, CVPR 2019)  & \XSolidBrush       & 37.9          \\
        MAF (\cite{maf}, CVPR 2019)         & \XSolidBrush       & 41.0          \\
        AT-Faster (\cite{at_faster}, ECCV 2020)   & \XSolidBrush       & 42.1          \\
        CST-DA (\cite{cst_da}, ECCV 2020)      & \XSolidBrush       & 43.6          \\ 
        KTNet (\cite{KTNET}, ICCV 2021)      & \XSolidBrush       & 45.6          \\ \midrule
        SFOD-Mosaic (\cite{free_lunch}, AAAI 2021) & \Checkmark       & 44.6          \\
        LODS (\cite{lods}, CVPR 2022)        & \Checkmark       & 43.9          \\
        $A^{2}SFOD$ (\cite{a2sfod}, AAAI 2023)      & \Checkmark       & 44.9          \\
        IRG (\cite{IRG}, CVPR 2023)      & \Checkmark       & 46.9          \\
        Ours        & \Checkmark       & \textbf{48.4} \\ 
        \bottomrule
    \end{tabular}
\end{table}

\begin{table}
  \caption{Experimental results of adaptation from synthetic to real images, \textit{i.e.}, from Sim10k to Cityscapes.}
  \label{tab:S2c}
    \begin{tabular}{@{}ccc@{}}
        \toprule
        Methods     & Source-free & AP of car     \\ \midrule
        Source only & \XSolidBrush       & 34.2          \\ \midrule
        DA-Faster (\cite{da_faster}, CVPR 2018)   & \XSolidBrush       & 38.5          \\
        SW-Faster (\cite{sw_faster}, CVPR 2019)   & \XSolidBrush       & 40.1          \\
        MAF (\cite{maf}, CVPR 2019)         & \XSolidBrush       & 41.1          \\
        AT-Faster (\cite{at_faster}, ECCV 2020)   & \XSolidBrush       & 42.1          \\
        Unbiased DA (\cite{unbiased_da}, CVPR 2021) & \XSolidBrush       & 43.1          \\
        KTNet (\cite{KTNET}, ICCV 2021)      & \XSolidBrush       & 50.7          \\ \midrule
        SFOD-Mosaic (\cite{free_lunch}, AAAI 2021) & \Checkmark       & 43.1          \\
        $A^{2}SFOD$ (\cite{a2sfod}, AAAI 2023)      & \Checkmark       & 44.0          \\
        IRG (\cite{IRG}, CVPR 2023)      & \Checkmark       & 45.2          \\
        Ours        & \Checkmark       & \textbf{47.3} \\ 
        \bottomrule
    \end{tabular}
\end{table}

\begin{table*}
  \caption{Experimental results of adaptation to a large-scale dataset, \textit{i.e.}, from Cityscapes to BDD100k daytime.}
  \label{tab:c2b}
    \begin{tabular}{@{}cccccccccc@{}}
        \toprule
        Methods     & Source-free   & truck         & car           & rider         & person        & motor         & bicycle       & bus           & mAP           \\ \midrule
        Source only  &\XSolidBrush   & 11.0    & 46.4          & 26.6          & 30.3          & 11.8          & 20.5          & 10.7           &  22.5             \\ \midrule
        DA-Faster (\cite{da_faster}, CVPR 2018)  &\XSolidBrush    & 14.3          & 44.6          & 26.5          & 29.4          & 15.8          & 20.6          & 16.8          & 24.0          \\
        SW-Faster (\cite{sw_faster}, CVPR 2019)  &\XSolidBrush    & 15.2          & 45.7          & 29.5          & 30.2          & 17.1          & 21.2          & 18.4          & 25.3          \\
        Categorical DA (\cite{categorical_da}, CVPR 2020) &\XSolidBrush  & 19.5          & 46.3          & 31.3          & 31.4          & 17.3          & 23.8          & 18.9          & 26.9          \\ 
        SFA (\cite{SFA}, ACM MM 2021) &\XSolidBrush  & 19.1          & 57.5          & 27.6          & 40.2          & 15.4          & 19.2          & 23.4          & 28.9          \\ \midrule
        SFOD-Mosaic (\cite{free_lunch}, AAAI 2021)  & \Checkmark   & 20.6          & 50.4          & 32.6          & 32.4          & 18.9          & 25.0          & 23.4          & 29.0          \\
        $A^{2}SFOD$ (\cite{a2sfod}, AAAI 2023)  & \Checkmark    & \textbf{26.6} & 50.2          & 36.3          & 33.2          & \textbf{22.5} & 28.2          & \textbf{24.4}          & 31.6          \\
        Ours   & \Checkmark        & 24.5          & \textbf{55.2} & \textbf{38.9} & \textbf{41.4} & 20.9          & \textbf{30.4} & 23.2 & \textbf{33.5} \\ 
        \bottomrule
    \end{tabular}
\end{table*}


\begin{table}[]
    \caption{Ablation studies of our proposed LPU on Cityscapes ${\rightarrow}$ Foggy-Cityscapes. MT denotes the mean teacher framework. PST and LSCL denote Proposal Soft Training and Local Spatial Contrastive Learning, respectively.}
    \centering
    \begin{tabular}{@{}cccc@{}}
    \toprule
    Methods  & PST & LSCL                & mAP       \\ \midrule
    MT ($\sigma_{h}=\sigma_{l}=0.8$)       &     &                 & 35.0       \\ \midrule
             &\Checkmark &               &37.3          \\
    Our Proposed &    &\Checkmark         &37.1           \\
             &\Checkmark    &\Checkmark    & \textbf{38.8} \\ \bottomrule
    \end{tabular}
    \label{tab:ablation}
\end{table}

\begin{table}[]
    \caption{Results of different thresholds on Cityscapes ${\rightarrow}$ Foggy-Cityscapes. LPU denotes Low-confidence Pseudo-labels Utilization module. }
    \centering
    \begin{tabular}{@{}ccc@{}}
    \toprule
    Threshold & LPU & mAP \\ \midrule
    $\sigma_{h}=\sigma_{l}=0.8$     &     &35.0     \\
    $\sigma_{h}=\sigma_{l}=0.1$     &     &32.0     \\
    $\sigma_{h}=1.0, \sigma_{l}=0.1$  &\Checkmark     &36.3     \\
    $\sigma_{h}=0.8, \sigma_{l}=0.1$  &\Checkmark    &38.8     \\ \bottomrule
    \end{tabular}
    \label{tab:theshold}
\end{table}

\subsection{Datasets}
We consider five datasets frequently used in the UDA and SFDA literatures to validate the proposed approach. (1) Cityscape \cite{cityscapes} contains numerous photographs of outdoor street scenes captured from various cities. It comprises 2,975 training images and 500 validation images and includes eight object categories:  person, rider, car, truck, bus, train, motorcycle, and bicycle. (2) Foggy-Cityscapes \cite{foggy_cityscape} utilizes the images from Cityscapes to simulate foggy conditions while inheriting the annotations of  Cityscapes. It consists of the same number of images as Cityscapes. (3) KITTI \cite{kitti} is a widely used autonomous driving dataset consisting of manually collected images captured in various urban scenes, with 7,481  labeled images available for training. (4) Sim10k \cite{sim10k} is a simulation dataset generated from the popular computer game  Grand Theft Auto V, comprising 10,000 synthetic driving scene images. (5) BDD100k \cite{bdd100k} is a dataset comprising 100,000 images captured in six scenes and under six weather conditions, with three categories indicating the time of day. In line with the previous work, we have extracted a subset of BDD100k labeled as "daytime," comprising 36,728  training and 5,258 validation images.
\begin{figure}[h]
    \centering
    \includegraphics{ 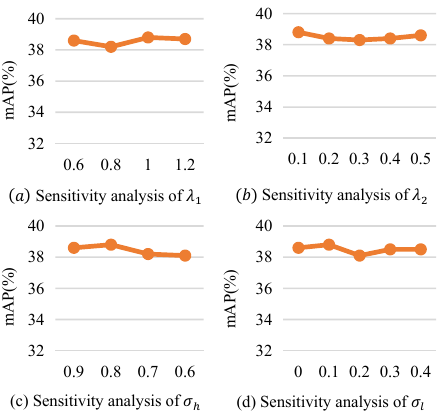}
    \caption{Hyperparameter analysis with respect to $\lambda_{1}$, $\lambda_{2}$, ${\sigma_h}$, and ${\sigma_l}$.}
    \label{fig:sensitivity}
\end{figure}

\subsection{Implementation Details}
Following the SFDA setting \cite{free_lunch,a2sfod}, we adopt FasterRCNN \cite{faster_rcnn} with ImageNet \cite{imagenet} pre-trained VGG-16 \cite{vgg} as the backbone. Both the source and student models are trained using SGD optimizer with a learning rate of 0.001 and momentum of 0.9. For the mean-teacher framework, the weight momentum update parameter $\alpha$ of the EMA for the teacher model is set equal to 0.996. $\lambda_{1}$ and $\lambda_{2}$ are set to 1 and 0.1 respectively. The temperature coefficient $\tau$ is set to 0.07. The high threshold $\sigma_{h}$ and the low threshold $\sigma_{l}$ are set to 0.8 and 0.1, respectively. During the evaluation, we report the mean Average Precision (mAP) with an IoU threshold of 0.5 on the target domain.

\subsection{Comparison with Existing SOTA Methods}
This section compares our proposed method with existing state-of-the-art UDAOD and SFOD methods. SFOD methods include SFOD-Mosaic \cite{free_lunch}, LODS \cite{lods}, $A^{2}SFOD$ \cite{a2sfod} and IRG \cite{IRG}. "Source Only"  denotes models trained on source domain data. The quantitative results cited in the tables for all the methods above are taken from their original papers. 

\subsubsection{C2F: Adaptation from Clear to Foggy Weather}
In many cases, deployed detectors are trained solely under clear weather conditions. However, in real-world scenarios like autonomous driving, these models may confront diverse weather conditions, including fog and haze. To evaluate the effectiveness of the proposed method in such conditions, Cityscapes are utilized as the source domain,  while Foggy Cityscapes serve as the target domain. Table \ref{tab:c2f} displays that our proposed method significantly outperforms all SFOD methods and improves the best-performing state-of-the-art IRG \cite{IRG} by +1.7\% mAP. In addition, our method can be comparable with some UDAOD methods, which can use source and target domain data. Our approach has achieved notable success in most categories of AP scores, including "car" from 51.9\% to 55.4\%, "rider" from 45.7\% to 50.3\%, "person" from 37.4\% to 39.0\%, "bicycle" from 41.6\% to 44.2\%, and "bus" from 39.7\% to 46.0\%. The achievement is due to our complete exploitation of low-confidence pseudo-labels, which is challenging to accomplish using the traditional single threshold setting.

\begin{table}
    \caption{Results of different strategies for mixup on Cityscapes ${\rightarrow}$ Foggy-Cityscapes. }
    \centering
    \begin{tabular}{@{}llll@{}}
    \toprule
    Mixup & mAP & Mixup & mAP \\ \midrule
    $random$      &36.9     &$random^{-}$       &36.6     \\
    $cls-mixup$      &38.2     &$cls-mixup^{-}$       &37.6     \\ 
    $IoU-mixup$      &38.8     &$IoU-mixup^{-}$       &38.1     \\ \bottomrule
    \end{tabular}
    \label{tab:mixup}
\end{table}

\subsubsection{K2C: Adaptation to A New Sense. }
In realistic scenes, it is common for cameras to have varying configurations, such as resolution and angle. This variation often leads to domain shift, which can affect the performance of the deployed detectors. To evaluate the capability of a model to adapt to unseen new scenes, the KITTI  and Cityscapes datasets are utilized as the source and target domains,  respectively. The experimental results are presented in Table \ref{tab:k2c}. Our proposed method achieves the highest AP under the SFDA setting.

\subsubsection{S2C: Adaptation from Synthetic to Real Images. }
Labeling detection data in autonomous driving is computationally intensive and time-consuming. As a result, it is reasonable to train detector models on synthetically generated datasets and subsequently deploy them in the real world. Nevertheless, the large domain gap between real and synthetic data hinders this deployment. In this experiment, we use Sim10k as the source domain and Cityscapes as the target domain. As shown in Table \ref{tab:S2c}, our method outperforms all SFOD methods, significantly exceeding the IRG \cite{IRG} by +2.1\% mAP. 

\subsubsection{C2B: Adaptation to Large-Scale Dataset. }
Although it is now easy to collect vast amounts of image data, labeling these data continues to pose a significant challenge for supervised learning methods. Therefore, ensuring the transferability of limited labeled data to unlabeled large-scale target datasets is crucial. In Table \ref{tab:c2b}, our method achieves 33.5\%  mAP, a +1.9\%  mAP improvement over the current best result.

\subsection{Further Analysis}
\subsubsection{Ablation Study. } 
An ablation study is conducted under the adaptation scenario of Cityscapes ${\rightarrow}$ Foggy-Cityscape to investigate the effectiveness of each module. The experimental results are in Table \ref{tab:ablation}. We assess the effectiveness of the LPU in leveraging low-confidence pseudo-labels. All experiments utilize the same threshold settings (\textit{i.e.}, $\sigma_{h}=0.8, \sigma_{l}=0.1$). With only the PST module, the model achieves 37.3\% mAP, demonstrating the effectiveness of PST in generating soft labels for soft training. In addition, the mAP improves to 37.1\% by LSCL alone, validating the correctness of our exploration of the stability of local spatial prediction. Eventually, with both PST and LSCL modules, the mAP increases to 38.8\%, indicating that the two components are functionally complementary and synergize to use low-confidence pseudo-labels effectively.

\subsubsection{Parameter Sensitivity Analysis. } 
This section investigates the sensitivity of hyperparameters on the transfer scenario Cityscapes ${\rightarrow}$ Foggy-Cityscape. The results are displayed in Figure \ref{fig:sensitivity}. We examine the sensitivity of hyperparameters $\lambda_{1}$ and $\lambda_{2}$, which control the weight of $\mathcal{L}_{pst}$ and $\mathcal{L}_{lscl}$ in Equation \ref{equ_10}. The sensitivity analysis is illustrated in Figure \ref{fig:sensitivity}(a) and Figure \ref{fig:sensitivity}(b). The method yields a relatively stable result across a wide range of $\lambda_{1}$ and $\lambda_{2}$. 

\subsubsection{Threshold Analysis. }
We investigate the effect of setting different confidence thresholds on the model. (1) We evaluate the performance of the mean-teacher framework as a baseline for training with a single threshold (\textit{i.e.}, $\sigma_{h}$=$\sigma_{l}$). As indicated by the first two experiments in Table \ref{tab:theshold}, the mean teacher framework achieves 35.0\% mAP when using a threshold of 0.8 and 32.0\% mAP when the threshold is set to 0.1. The reduced performance with the latter threshold is attributed to less accurate pseudo-label information resulting from a lower confidence threshold, which misles the model training. 
(2) We examine the necessity of distinguishing between high- and low-confidence pseudo-labels. Setting $\sigma_{h}$ to 1.0 and $\sigma_{l}$ to 0.1, whereby all pseudo-labels with confidence scores greater than 0.1 are leveraged using the LPU method, yielded a 36.3\% mAP performance by the model. This represents a 4.3\% mAP improvement over the conventional approach (\textit{i.e.}, $\sigma_{h}$=$\sigma_{l}=0.1$), demonstrating the superiority of LPU in leveraging low-confidence pseudo-labels. Nonetheless, this approach is not as effective as treating high- and low-confidence pseudo-labels separately (\textit{i.e.}, $\sigma_{h}=0.8, \sigma_{l}=0.1$). Since high-confidence pseudo labels are trustworthy enough to be trained with greater accuracy and efficiency using the traditional method.
(3) We investigate the sensitivity of setting different confidence thresholds. The results are presented in Figure \ref{fig:sensitivity}(c) and Figure \ref{fig:sensitivity}(d), which display a range of thresholds. The results confirm that our method maintains good performance across a wide range of thresholds, effectively demonstrating the usefulness of LPU for training using pseudo-labels.

\subsubsection{Mixup Analysis. } 
To evaluate the effectiveness of the LSCL module in enhancing the model's ability to distinguish spatially neighboring proposals, we replace IoU-mixup with two other approaches: random mixup and cls-mixup. Random mixup refers to randomly selecting proposals from $\left\{\boldsymbol{T}_{n}\right\}_{n=1}^{N_{p}}$ for mixup. As for cls-mixup, the proposals used for mixup are selected based on the similarity of their output probability vectors after passing through the RCNN classification layer. The experimental results are shown in Table \ref{tab:mixup}. The two columns on the left indicate that the features participating in the mixup are from the student and teacher, respectively. In comparison, the two columns on the right with the superscript "${^{-}}$" indicate that all the features participating in the mixup are from the student. We can make the following observations: (1) As we mentioned before, the performance of the mixup using the features generated by the student and teacher is better. (2) Random mixup does not significantly enhance the model's performance. At the same time, cls-mixup improves the mAP score to 38.2\%, indicating that the proposals can be further optimized using category relationships. However, the improvement is limited because the category information of these proposals is not very accurate. (3) IoU-mixup improves the mAP score to 38.8\%, highlighting the importance of IoU-mixup and verifying the efficacy of the LSCL module in leveraging local spatial relationships for optimization purposes. To better understand how the LSCL module improves model performance, please see the Appendix.

\section{CONCLUSION}
We propose a novel paradigm for source-free object detection (SFOD), which efficiently utilizes low-confidence pseudo-labels by introducing the  Low-confidence Pseudo-labels Utilization (LPU) module. The LPU module consists of Proposal Soft Training (PST) and Local Spatial Contrastive Learning (LSCL). PST generates soft labels for proposals, while LSCL optimizes the representational features by exploiting local spatial relationships. To demonstrate the effectiveness of our proposed method, we perform extensive experiments on five cross-domain object detection datasets. The results of our experiments demonstrate that our approach surpasses the performance of the state-of-the-art source-free domain adaptation and many unsupervised domain adaptation methods.

\begin{acks}
This work is supported by the National Natural Science Foundation of China under Grant 62176246 and Grant 61836008. This work is also supported by Anhui Province Key Research and Development Plan (202304a05020045), Anhui Province Natural Science Foundation (2208085UD17), and the Fundamental Research Funds for the Central Universities (WK3490000006).
\end{acks}

\bibliographystyle{ACM-Reference-Format}
\balance
\bibliography{sample-sigconf}


\appendix

\section{Analysis of the LSCL module}

\begin{figure}[h]
  \centering
  \includegraphics[width=\linewidth]{ 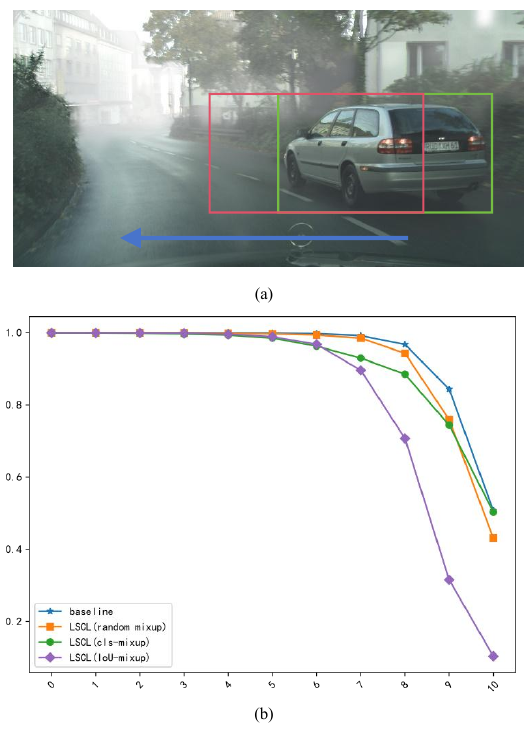}
  \caption{An experiment to illustrate how the LSCL module (especially the IoU-mixup operation) improves the model performance} \label{fig:lscl_rb}
\end{figure}

We conduct an experiment to explain how the LSCL module, particularly the IoU-mixup operation, enhances the model's performance.
In Figure \ref{fig:lscl_rb}(a), the green bbox represents the ground truth, and the IoU between the green and red bboxes is 0.5. We performed a gradual horizontal movement of a bounding box, starting from the green bbox and ending at the red bbox. This movement was divided into ten steps, and at each step, we input the generated bbox into the model, obtained the classification probability, recorded the highest category probability, and plotted its change curve in Figure \ref{fig:lscl_rb}(b).

LSCL module employs contrastive loss on the features produced after IoU-mixup to learn the proximity of adjacent proposals and identify the relative similarity between neighboring and other proposals. Through these experiments, we observed that adding the LSCL module, especially using IoU-mixup, makes the model more sensitive to changes in bbox positions. Consequently, the model is encouraged to explore finer-grained cues between neighboring proposals during training, forming more robust classification boundaries. Simultaneously, the model is prompted to optimize for bboxes with more accurate positions (as depicted in Figure \ref{fig:lscl_rb}(b), the bboxes corresponding to LSCL, especially when using IoU-mixup, are closer to the ground truth for the same confidence score).

\end{document}